\documentclass[preprint,5p,twocolumn]{elsarticle}

\usepackage{framed,multirow}

\usepackage{url}
\usepackage{xcolor}
\definecolor{newcolor}{rgb}{.8,.349,.1}

\usepackage{geometry} 
\usepackage{graphicx}
\usepackage{subcaption}
\usepackage{caption}
\usepackage{hhline}
\usepackage{siunitx}
\usepackage{amsmath}
\usepackage{amssymb}
\usepackage{multirow}
\usepackage{blindtext, xcolor}
\usepackage{diagbox}
\usepackage{tabularx}
\usepackage{bookmark}
\hypersetup{
  urlcolor=Black, linkcolor=Black, citecolor=Black, 
}

\begin{document}
\begin{frontmatter}
\title{Active Learning for Segmentation Based on Bayesian Sample Queries} 

\author[ETH]{Firat Ozdemir\corref{cor1}}
\ead{ozdemirf@ethz.ch}
\author[ETH]{Zixuan Peng}
\author[Balgrist]{Philipp Fuernstahl}
\author[ETH]{Christine Tanner}
\author[ETH]{Orcun Goksel}

\cortext[cor1]{Corresponding author.}
\address[ETH]{Computer-assisted Applications in Medicine, ETH Zurich, Switzerland}
\address[Balgrist]{Computer Assisted Research and Development, University Hospital Balgrist, University of Zurich, Switzerland}

\begin{abstract}
Segmentation of anatomical structures is a fundamental image analysis task for many applications in the medical field. 
Deep learning methods have been shown to perform well, but for this purpose large numbers of manual annotations are needed in the first place, which necessitate prohibitive levels of resources that are often unavailable.
In an active learning framework of selecting informed samples for manual labeling, expert clinician time for manual annotation can be optimally utilized, enabling the establishment of large labeled datasets for machine learning.
In this paper, we propose a novel method that combines \emph{representativeness} with \emph{uncertainty} in order to estimate ideal samples to be annotated, iteratively from a given dataset.
Our novel representativeness metric is based on Bayesian sampling, by using information-maximizing autoencoders.
We conduct experiments on a shoulder magnetic resonance imaging (MRI) dataset for the segmentation of four musculoskeletal tissue classes.
Quantitative results show that the annotation of representative samples selected by our proposed querying method yields an improved segmentation performance at each active learning iteration, compared to a baseline method that also employs uncertainty and representativeness metrics.
For instance, with only $10\%$ of the dataset annotated, our method reaches within $5\%$ of Dice score expected from the upper bound scenario of all the dataset given as annotated (an impractical scenario due to resource constraints), and this gap drops down to a mere $2\%$ when less than a fifth of the dataset samples are annotated.
Such active learning approach to selecting samples to annotate enables an optimal use of the expert clinician time, being often the bottleneck in realizing machine learning solutions in medicine.
\end{abstract}

\end{frontmatter}

\section{Introduction}
\label{sec:introduction}

Technological advancements allow for both longer life expectancy and higher quality of life. 
These both increase demand on medical personnel, who are also expected more and more to perform personalized and patient-specific procedures, such as surgical planning via morphological approaches~\cite{furnstahl2016surgical} or functional simulation~\cite{pean2017physical}. 
To that end, even when it is possible to see target anatomical structures in an imaging modality such as MRI, CT, or ultrasound, it is still often the bottleneck to automatically identify and delineate (segment) them. 
Due to limited resources for manual annotations, patient-specific procedures are still not a common practice for most clinical applications.

In the recent years, deep learning (DL) has shown encouraging performance for segmentation when sufficient amount of annotated data for the anatomical structure of interest is available.
Annotating a sufficiently large dataset by medical experts is a time- and hence cost-intensive undertaking. 
The idea of \textit{active learning} is to identify the samples that, once annotated, will bring the most value, which can be defined, e.g., as the gain in segmentation performance of the learned model.
In an iterative process, the developed framework selects a new set of samples --~also referred to as \textit{batch-mode} active learning~-- to be manually annotated at each active learning iteration. 
This is inherently feasible in the clinical environment, where medical experts anyhow annotate small batches of images at different intervals based on their availability between daily clinical responsibilities. 
In a clinical setting, typically at each annotation session, image data to be annotated is loaded from a picture archiving and communication system (PACS). 
A \textit{pool-based} active learning system can thus intervene at that stage, in order to intelligently determine which volumes or which image slices to display and request the user to annotate.

Active learning with DL remains a challenging problem, since DL solutions do not typically generalize well to unseen samples. 
Hence, there have been a wide range of approaches in the literature to improve sample selection in active learning.
Most of these works can be grouped under \textit{uncertainty} and \textit{representation} based sampling methodologies.
\\
\noindent\textbf{Uncertainty Sampling.}
In~\cite{gal2017deep,gal2016dropout}, it was shown that \emph{dropout} layers can be used at inference time to sample from the approximate posterior, so-called Monte Carlo (MC) Dropouts.
This gives flexibility for sampling as many posteriors as desired, with virtually zero cost added during training; i.e., a similar cost for training a single model as opposed to an ensemble.
Then, the disagreement among posteriors, e.g.,\, variance, can be used to quantify the uncertainty.
In~\cite{matthias2018deep} a classificiation approach was proposed, where ``pseudo-labels'' are assigned on non-annotated samples using a network trained on a small annotated sample set. 
The objective at an active learning iteration is then to keep prediction accuracy as high as possible on the annotated sample set while using MC Dropouts to query for the most uncertain non-annotated samples to be annotated. 
In~\cite{konyushkova2019geometry}, the proposed method queries a patch from 3D volumes using a combination of geometric smoothness priors and entropy-based novel uncertainty measures.
\\
\noindent\textbf{Representation Sampling.}
Uncertainty quantification with DL models can lead to out-of-distribution samples being ignored in the active learning process~\cite{sener2017geometric}. 
Consequently, population coverage for active learning is widely investigated.
In~\cite{sener2017active}, the authors propose a greedy sample selection algorithm using the last fully connected layer of a Convolutional Neural Network (CNN) to solve maximum set-cover~\cite{Feige1998a} between the pool of all images and the unison of currently annotated samples and the next sample to be queried.
In~\cite{yang2017suggestive}, a similar representation sampling method is coupled with uncertainty sampling when tackling active learning for semantic segmentation.
The authors compute uncertainty measure as the variance of predictions from multiple CNNs, where each CNN is trained with a bootstrap of the available dataset.
Next, a representative subset of the most uncertain samples are sought by computing the angle between \textit{image descriptor} vectors $x^\mathrm{id}$, defined as the spatially averaged activation tensor from the CNN where the spatial resolution is the coarsest. 
Distance metric approaches in high-dimensional spaces suffer from the so-called\textit{distance concentration}~\cite{franccois2008high}, which is a limitation of both works~\cite{sener2017active} and \cite{yang2017suggestive} above.

Note that with the methods described above, the not-yet-annotated dataset is only weakly integrated at any stage prior to quantifying a fitness metric of samples from that dataset. 
In other words, a posterior estimated from the relatively small annotated dataset is taken to be a good predictor of the complete dataset distribution. 
This is a strong assumption, especially at early active learning iterations when the annotated set size is still small. 
Powerful tools are proposed for unsupervised DL, such as Autoencoders (AEs)~\cite{bengio2007greedy}, which learn to map (encode) the high dimensional input space onto a manifold of substantially lower dimensions, such that it can reconstruct the high dimensional input with a second mapping function (decoder).
Variational autoencoders (VAEs)~\cite{Kingma2013autoencoding} build on AEs, with additional regularization enforced in a latent space. 
This regularization constraint penalizes the encoder part of the network such that the training dataset is mapped onto a known prior distribution of some random variables, often modelled as standard normal distribution. 
Intuitively, this regularization promotes the creation of a continuous latent space of the observed samples.
In ideal conditions, this means that traversing the manifold from the latent vector of one image to another, one can generate realistic samples that change from the former image to the latter.
In active learning, having such an embedding space explaining the non-annotated data is a formidable source of information that can readily be exploited in order to ensure that key samples from the given population are queried for annotation early on.
In~\cite{zhao2017infovae} the authors show that latent space of VAE can be suboptimal, hence they propose the method \textit{infoVAE}, which uses Maximum-Mean Discrepancy (MMD)~\cite{Gretton2006a,Li2015generative} instead of the KL-divergence measure as MMD learns a more continuous and informative latent space representation. 
Recently the authors of~\cite{sinha2019variational} presented an active learning framework where they train a VAE on all available images in an adversarial fashion with a discriminator classifying between annotated and non-annotated samples.
Then, sample selection can be done with the discriminator using the latent space of the trained VAE, potentially solving the distance concentration problem in high-dimensionality of earlier works.

In the medical field, UNet~\cite{ronneberger2015unet} and DCAN~\cite{Chen2016DCAN} are some of the most popular neural network architectures for segmentation.
Most work in the field of medical image analysis have adopted the UNet approach, thanks to its intuitive structure and consistently high performance in pixel-level tasks, such as~\cite{Zeng20173dunet,milletari2016vnet,ozdemir2018learn,Salehi2017precise}.
On the other hand, DCAN won the 2015 MICCAI Gland Segmentation Challenge~\cite{Sirinukunwattana2017gland}.
Thanks to its deeply supervised~\cite{lee2015deeply,guo2019btsdsn} architecture, DCAN can be trained faster, thereby being particularly attractive in active learning~\cite{yang2017suggestive}.

In earlier work~\cite{ozdemir2018active}, we achieved state of the art results in active learning for the segmentation of a shoulder MR dataset.
Inspired by~\cite{yang2017suggestive}, we proposed to have metrics quantifying both uncertainty and representativeness for selecting the next batch of samples. 
In contrast to~\cite{yang2017suggestive}, we used variance from MC Dropout samples~\cite{gal2016dropout} as an uncertainty metric, experimented with different representativeness metrics, explored different means to combine uncertainty and representativeness measures, and proposed a latent space regularization term that promotes maximizing its information content during training of the segmentation network.
Although the optimization of maximum entropy in the latent space can be counter-intuitive for segmentation, our results in~\cite{ozdemir2018active} showed that it can help ensure to generate a discriminative representation of the image dataset.

In this work, we approach the representativeness measure from a probabilistic point of view, where we optimize for the MMD~\cite{Gretton2006a} divergence using VAEs to learn meaningful latent features which follow a Gaussian distribution.
This is herein studied for a segmentation task using a Bayesian approach for an efficient coverage of the entire set of images with the significantly smaller set of annotated images.
Our representation sampling is agnostic to the current and future tasks, i.e.,\, independent of the task. 
Similarly to~\cite{sinha2019variational}, we herein adopt the idea of VAEs for a low dimensional representation for sampling.
Additionally, we herein incorporate an uncertainty-based sampling criterion to further promote relevant sample selection. 
We utilize VAEs particularly with MMD, which was shown in~\cite{zhao2017infovae} to improve latent space representations. 
Note that for our purposes, in contrast to earlier works, an additional training for representation sampling is not needed at new active learning iterations, which is an important advantage since the pool of medical image datasets can be vast and prohibitive for regular additional training in the clinical setting.

\section{Methods}
\label{sec:methods}

\subsection{Notation}
Below we define the notations used in this manuscript. 

\noindent
\textbf{Dataset.}
Let the pool of all images $D_\mathrm{pool}$ consist of images $\{x\}$ and their annotations $\{y\}$, the latter of which in an active learning iteration would be partially inaccessible for images not yet annotated.
At a given active learning iteration $t$, there would then be a readily annotated dataset $D_\mathrm{an}^{(t)} \subset D_\mathrm{pool}$.
The not-yet-annotated dataset is referred to as $D_\mathrm{non}^{(t)} = D_\mathrm{pool}^{(t)} \setminus D_\mathrm{an}^{(t)}$, which in practice have only images available.
For brevity, we will omit the active learning iteration representation $(\cdot)^{(t)}$ for descriptions within an iteration and use this only for formulations that affect multiple iterations.
Note that typically $|D_\mathrm{an}| \ll |D_\mathrm{pool}|$, since active learning would be redundant if the set sizes were of similar cardinality.
We will treat these sets as random variables, hence observations from the annotated and the pool image sets then become $x_\mathrm{an} \sim X_\mathrm{an}$ and $x_\mathrm{pool} \sim X_\mathrm{pool}$, respectively.
At an active learning iteration, i.e.,\, prior to each manual annotation session, a method should select a set of samples $S_\mathrm{query}$ to be annotated, where $S_\mathrm{query} \subset D_\mathrm{non}$.
Once annotated by the user, these samples will be appended to the annotated dataset along with their manual annotations, yielding $D_\mathrm{an}^{(t+1)}$ for the next iteration of active learning.

\vspace{1ex}\noindent
\textbf{Architecture.}
The architecture of our fully convolutional networks (FCNs) for segmentation follows a DCAN-like structure~\cite{yang2017suggestive}, where the receptive field of the convolutional kernels increase through max-pooling operation, creating spatially coarser feature maps while increasing the number of feature channels being learned. 
We call the spatially coarsest level of the network as \textit{abstraction layer}~\cite{ozdemir2018active}, which is relevant for the baseline method we will be comparing against.
Segmentation models are trained using pairs of images and annotations $\{x_i, y_i\} \in D_\mathrm{an}$.
For all VAE-based methods, the learned embedding space $Z \in \mathbb{R}^{n_\mathrm{lat}}$ is defined by $n_\mathrm{lat}$ latent variables.
VAE models are trained using only images $\{x_i\} \in D_\mathrm{pool}$.
Without loss of generality different network architectures can also be envisioned for our active learning approach proposed in this work. 
What is essential is to accommodate necessary modules in the segmentation model to be able to quantify uncertainty, and estimate a latent space that can represent the image population in the form of a normal distribution for representativeness quantification.

\subsection{Quantifying Uncertainty}
\label{sec:uncertainty}

Model uncertainty expected from segmenting a non-annotated image is undoubtedly one of the most important cues to aim for in active learning. 
However, uncertainty is not inherently quantified in most CNNs.
Consider a conventional supervised segmentation task using dataset $D_\mathrm{an}$.
For an observation $x$, the task can be formulated as computing the maximum a posteriori $p(y|x, \Theta)$, where $\Theta$ is the set of learned model parameters using $X_\mathrm{an}$ and $Y_\mathrm{an}$.
This can be formulated as
\begin{equation}
    \begin{split}
    y^{*} & = \arg\max_{y} \int p(y|x, \theta) p(\theta|X_\mathrm{an}, Y_\mathrm{an}) d\theta \\
    & \approx p(y|x, \Theta) \text{ s.t. } \Theta = \arg\max_{\theta} p(\theta|X_\mathrm{an}, Y_\mathrm{an}) \text{, } 
    \end{split}    
\label{eqn:segmentation_formula}
\end{equation}
where the maximum a posteriori for $\theta$ is instead learned due to the impracticalities for integrating over high dimensional $\theta$.
This then leads to deterministic predictions for $y$.
In order to approximate $p(y|x, \theta)$, MC Dropouts is proposed in~\cite{gal2017deep} to sample from model parameters, aggregating desired number of posterior predictions with only additional inference operations. 

In order to leverage the benefits of MC Dropout, we modify the DCAN architecture~\cite{yang2017suggestive} with additional spatial dropout layers~\cite{tompson2015efficient}, similarly to~\cite{ozdemir2018active}.
First, we infer a tensor of segmentation predictions $p(y \!\! = \!\! l\, |\, x, \hat{\theta}) \in \mathbb{R}^{n_\mathrm{MC}, \mathrm{N}}$ for label $l$ given each draw of model parameters $\hat{\theta}$ depending on the random dropouts, where $n_\mathrm{MC}$ is the number of MC Dropout samples, and $N$ is the number of the input image pixels. 
Next, we compute the uncertainty map for label $l$ as the variance of each pixel prediction over $n_\mathrm{MC}$ inferences.
Finally, we compute a scalar uncertainty measure as the spatial average of this uncertainty map, yielding
\begin{equation}
    m_\mathrm{unc}(y=l) = \frac{1}{N}\sum_{n=1}^{N} \mathrm{var}\left[p(y=l\, |\,x, \hat{\theta})^{(n)}\right]
    \label{eqn:uncertainty}
\end{equation}
where $p(y \!\! = \!\! l\, |\, x, \hat{\theta})^{(n)}$ is the vector of $n_\mathrm{MC}$ predictions at pixel $n$.
In the multi-class setting where each anatomy is similarly important, we estimate the model uncertainty for the segmentation task as the mean of scalar uncertainty measures for each segmentation label.

\subsection{Maximum Likelihood Sampling in Latent Space}
\label{sec:latent_likelihood}

Note that the above quantification of model uncertainty for an observation $x_i$ is conditioned on the annotated dataset $D_\mathrm{an}$ but not on $D_\mathrm{pool}$. The latter is ideally needed for a good sample prediction for the image population.
Below, we describe an approach to take into account the potential domain shift from the already-annotated to the entire dataset using unsupervised learning.

The goal is to populate an image set $X_\mathrm{an}$ such that it provides a sufficiently good representative summary of $X_\mathrm{pool}$.
For this purpose, consider a mapping function ${f_\mathrm{enc}:x_i \mapsto z_i}$, where each observation $x_i$ from $X_\mathrm{pool}$ is mapped onto $Z \in \mathbb{R}^{n_\mathrm{lat}}$, a continuously defined latent space with a desired probability distribution, e.g.,\, a multivariate normal.

Intuitively, a batch of new queries for manual annotation from $X_\mathrm{non}$ after an active learning iteration $t$ should represent the distribution statistics of $X_\mathrm{pool}$ with an emphasis on the space that is unlikely for the distribution of $X_\mathrm{an}$.
In other words, queried samples $x_i$ should not be redundant due to readily existing samples in $X_\mathrm{an}^{(t-1)}$.
Provided that the mode of the latent space $Z$ will encode the most frequent attributes of $X_\mathrm{pool}$, the ideal sample $x^{*}$ can be queried as
\begin{equation}
  x^{*} =  \arg \!\!\!\!\! \max_{x_i\in X_\mathrm{non}} \frac{p(z|x_i, X_\mathrm{pool})}{p(z|x_i, X_\mathrm{an})} \ .
\label{eqn:goal_in_real_life}
\end{equation}

Over iterations, samples queried based on $x^{*}$ will align the posteriors $p(z|X_\mathrm{an})$ and $p(z|X_\mathrm{pool})$, making representations of observations from $X_\mathrm{an}$ cover both breadth and mode of $X_\mathrm{pool}$, hence achieving the desired objective.
To compute Eq.\,(\ref{eqn:goal_in_real_life}), we utilize Bayesian inference as 
\begin{equation}
	p(z|x_i, X) = \frac{p(x_i, X|z)p(z)}{p(x_i, X)} = \frac{p(x_i|X,z)p(X|z)p(z)}{p(x_i|X)p(X)} \text{ .} 
\label{eqn:bayes_step1}
\end{equation}
The right hand side contains the equivalent of the posterior $p(z|X)$, allowing for a simpler representation as,
\begin{equation}
	p(z|x_i, X) = \frac{p(x_i|X,z)p(z|X)}{p(x_i|X)} \propto p(x_i|X,z)p(z|X) \text{ .} 
\label{eqn:bayes_step2}
\end{equation}

In order to approximate $f_\mathrm{enc}$, we train an infoVAE~\cite{zhao2017infovae} with the complete pool of images $X_\mathrm{pool}$ using MMD for latent space regularization as $L_\mathrm{infoVAE} = L_\mathrm{AE} + L_\mathrm{MMD}$, where
\begin{equation}
\begin{split}
    L_\mathrm{MMD}(q||p) = & \mathbb{E}_{z\sim q, z' \sim q}[k(z,z')] + \mathbb{E}_{z\sim p, z' \sim p}[k(z,z')]\\
    &  - 2 \mathbb{E}_{z\sim q, z' \sim p}[k(z,z')] \text{ ,}
    \end{split}  
\end{equation}
$p$ is the prior, $q$ is the posterior inference in the latent space via the encoder, and $k(z,z')$ is the distance in a kernel space. 
We choose $p(z)$ to have a standard normal distribution and use a Gaussian as the kernel mapping $k(z,z') = \exp(-||z-z'||/2\sigma^2)$, where $\sigma$$=$$1$. 
Thereon, the true posterior inference $p(z|x)$ is approximated with $q_\phi(z|x)$, where $\phi$ is the learned parameter set of the infoVAE encoder. 
Hence, we can approximate Eq.\,(\ref{eqn:goal_in_real_life}) as
\begin{equation}
x^{*} \approx \arg \!\!\!\!\! \max_{x_i\in X_\mathrm{non}} \!\!\! [ \log(q_\phi(z|x_i, X_\mathrm{pool})) - \log(q_\phi(z|x_i, X_\mathrm{an}))] 
\label{eqn:log_bayes_query_numeric}
\end{equation}
and Eq.\,(\ref{eqn:bayes_step2}) as $q_\phi(x_i|X,z)q_\phi(z|X)$.
Accordingly, we project samples from (i) $X_\mathrm{an}$ and (ii) $X_\mathrm{pool}$ onto the latent space of the infoVAE.
Next, to compute $q_\phi(z|X)$, we fit a multivariate diagonal Gaussian to both projections, separately. 
Finally, we estimate the likelihoods $q_\phi(z|x_i, X_\mathrm{an})$ and $q_\phi(z|x_i, X_\mathrm{pool})$ using the error function $\mathrm{erf}(x)=\frac{1}{\sqrt{\pi}} \int_{-x}^{x}\exp{(-t^2)}dt$, as follows:
\begin{equation}
    q_\phi(z|x, X) \approx 1 + \mathrm{erf}\left(-\frac{|x - \mu_{X}|}{\sigma_{X}\sqrt{2}}\right) \text{ , }
\end{equation}
where $\mu_{X}$ and $\sigma_{X}$ are the parameters of the fitted Gaussians.
In other words, we use the first half of the cumulative distribution function of the fitted Gaussian since it is symmetric around its expected value $\mu_{X}$.

\begin{figure}
\centering
	\begin{subfigure}[b]{0.999\linewidth}
    \centering
        \includegraphics[width=0.999\linewidth, trim= 0cm 0.0cm 0cm 0cm, clip]{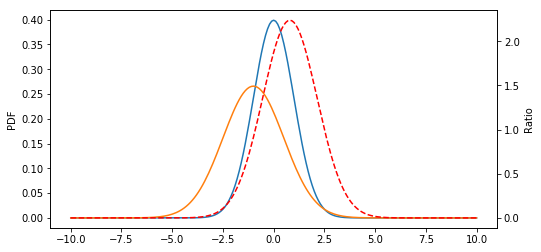}
    \end{subfigure}%
   	\caption{Toy example for Bayesian sample querying. 
   	Solid curves: Probability density function (pdf) of $q_\phi(z|X_\mathrm{pool})$ (blue) with $\mu$$=$$0$, $\sigma$$=$$1$ along with pdf of $q_\phi(z|X_\mathrm{an})$ with hypothetical $\mu$$=$$-1$, $\sigma$$=$$1.5$ (orange). 
   	Dashed red curve: non-normalized pdf of underrepresented samples in $X_\mathrm{an}$ given by ratio $q_\phi(z|X_\mathrm{pool}) / q_\phi(z|X_\mathrm{an})$.}
    \label{fig:bayesian_querying}
\end{figure}

Fig.~\ref{fig:bayesian_querying} illustrates this for a toy example, where $x^{*}$ would be selected based on maximizing the likelihood of being sampled from the non-normalized distribution shown with the dashed red Gaussian.
Additional experiments corroborating our intuition are provided in the Appendix.

\subsection{Comparative Evaluation}

We define 5 methods for analysis and comparison: 

\noindent
$\rightarrow \mathrm{FCN}_\mathrm{Random}$; a simplistic baseline approach of randomly selecting the samples to annotate; i.e.,\, random querying of $n_\mathrm{rep}$ samples.

\noindent
$\rightarrow \mathrm{FCN}_\mathrm{Uncertainty}$; the most uncertain $n_\mathrm{rep}$ samples based on Sec.~\ref{sec:uncertainty} are queried in each active learning iteration.

\noindent
$\rightarrow \mathrm{FCN}_\mathrm{Baseline}$; a baseline similar to~\cite{yang2017suggestive}, with the main difference being additional spatial dropout layers in the architecture, and using the uncertainty metric described in Sec.~\ref{sec:uncertainty} (instead of training 3 FCNs with different bootstrapped subsets of the available $D_\mathrm{an}$ and using variance across FCNs as in~\cite{ozdemir2018active}).
Consequently, the computational cost is reduced by a third and the entire $D_\mathrm{an}$ is observed by the trained model.
To be precise, first a set $S_\mathrm{unc}$ of the most uncertain $n_\mathrm{unc}$ elements from the non-annotated dataset $X_\mathrm{non}$ are selected. 
Next, \textit{image descriptor} $x^\mathrm{id} \in \mathbb{R}^{n_\mathrm{abs}}$ of each sample in $X_\mathrm{non}$ is computed as the global average pooling applied at the coarsest layer activations, where $n_\mathrm{abs}$ is the number of feature channels of the corresponding layer.
The representativeness metric can then be computed using the following similarity measure 
\begin{equation}
d_\mathrm{sim}(x_i, x_j) = \cos\left(x_i^\mathrm{id}, x_j^\mathrm{id}\right)
\end{equation}
between the two $x^\mathrm{id}$ vectors for any two images $x_i$ and $x_j$.
In an iterative manner, we populate a representative sample set $S_\mathrm{rep} \subset S_\mathrm{unc}$ by adding the currently most representative sample $x^*_\mathrm{rep}$ via~\cite{yang2017suggestive}
\begin{equation}
    x^*_\mathrm{rep} = \arg \max_{x_j \in S_\mathrm{unc} \setminus S_\textrm{rep}} \sum_{x_i \in X_\mathrm{non}} d_{\mathrm{sim}}(x_i, x_j \cup S_\textrm{rep}) \text{  .}
    \label{eqn:representativeness_metric}
\end{equation}
This maximizes maximum set-cover~\cite{Feige1998a} on $X_\mathrm{non}$ based on the $d_\mathrm{sim}$ metric.

\noindent
$\rightarrow \mathrm{FCN}_\mathrm{BSQ}$; \textit{Bayesian sample querying}, our proposed method, selects samples to be annotated based on the intersection of the most uncertain (Sec.~\ref{sec:uncertainty}) and representative samples following Eq.\,(\ref{eqn:log_bayes_query_numeric}).
Specifically, we first select the most uncertain samples from $X_\mathrm{non}$. 
Then, we form $S_\mathrm{rep} \subset S_\mathrm{unc}$ following Eq.\,(\ref{eqn:log_bayes_query_numeric}) with $n_\mathrm{rep}$ samples to be queried for annotation for the next active learning iteration.

\noindent
$\rightarrow \mathrm{FCN}_\mathrm{Upperbound}$; the upper bound using as a reference in our quantitative analysis. 
The upper bound uses the same segmentation architecture as the above compared methods, but is trained on the complete $D_\mathrm{pool}$ in a supervised setting; i.e.,\, assuming we already know all annotations at each sample query iteration.

\subsection{Implementation}
\label{sec:implementation}

For all compared methods, we used a modified DCAN architecture~\cite{ozdemir2018active} trained on 2D image input for the segmentation network using inverse frequency weighted cross entropy loss.
Data augmentation of horizontal flip was randomly applied to images with 0.5 uniform probability during training. 
When training, Adam optimizer was used with a learning rate of $5\times 10^{-4}$ and a mini-batch size of 8 images.
For both training and inference, dropout rate was set to 0.5, with $n_\mathrm{MC}$$=$$17$ for MC samples. 
At each active learning iteration including the initial training, models were trained for 8000 steps. 
We trained an infoVAE with 5 convolutional blocks in both encoder and decoder on downsampled images of size $96\times 96$, and we assigned dimensionality of the latent space as $n_\mathrm{lat}$$=$$200$.
For infoVAE training, Adam optimizer with learning rate of $5 \times 10^{-5}$ and a mini-batch size of 32 images was used.
Preprocessing of image-wise normalization was applied for the infoVAE training.
We used $l_2$-norm for the reconstruction loss $L_\mathrm{AE}$.
The methods were implemented and tested with the Tensorflow library on a cluster of NVIDIA Titan X GPUs.

\section{Experiments}
\label{sec:Experiments}

\begin{table}
\centering
\caption{Dataset consists of 2 different acquisition settings, which are merged together as shown in the 3rd row.
The depth resolution $\mathfrak{D}$ is 64 or 56, depending on the acquisition setting.}
\small
\label{tbl:dataset}
\setlength\tabcolsep{5.pt}
\begin{tabular}{r|r|r|r}
\multicolumn{1}{c|}{Setting} & \multicolumn{1}{c|}{\#volumes} & \multicolumn{1}{c|}{vox res.\,{[}mm{]}} & \multicolumn{1}{c}{digital res.\,{[}px{]}}   \\ \hline
\#1 & 20 & 0.91 x 0.91 x 3.0 & 192 x 192 x 64 \\ 
\#2 & 16 & 0.83 x 0.83 x 3.0 & 144 x 144 x 56  \\ \hline
Total & 36 & 0.91 x 0.91 x 3.0 & 192 x 192 x $\mathfrak{D}$ 
\end{tabular}
\end{table}

\begin{figure}
\centering
	\begin{subfigure}[b]{0.49\linewidth}
    \centering
        \includegraphics[width=0.999\textwidth, trim= 0cm 0.0cm 0cm 0cm, clip]{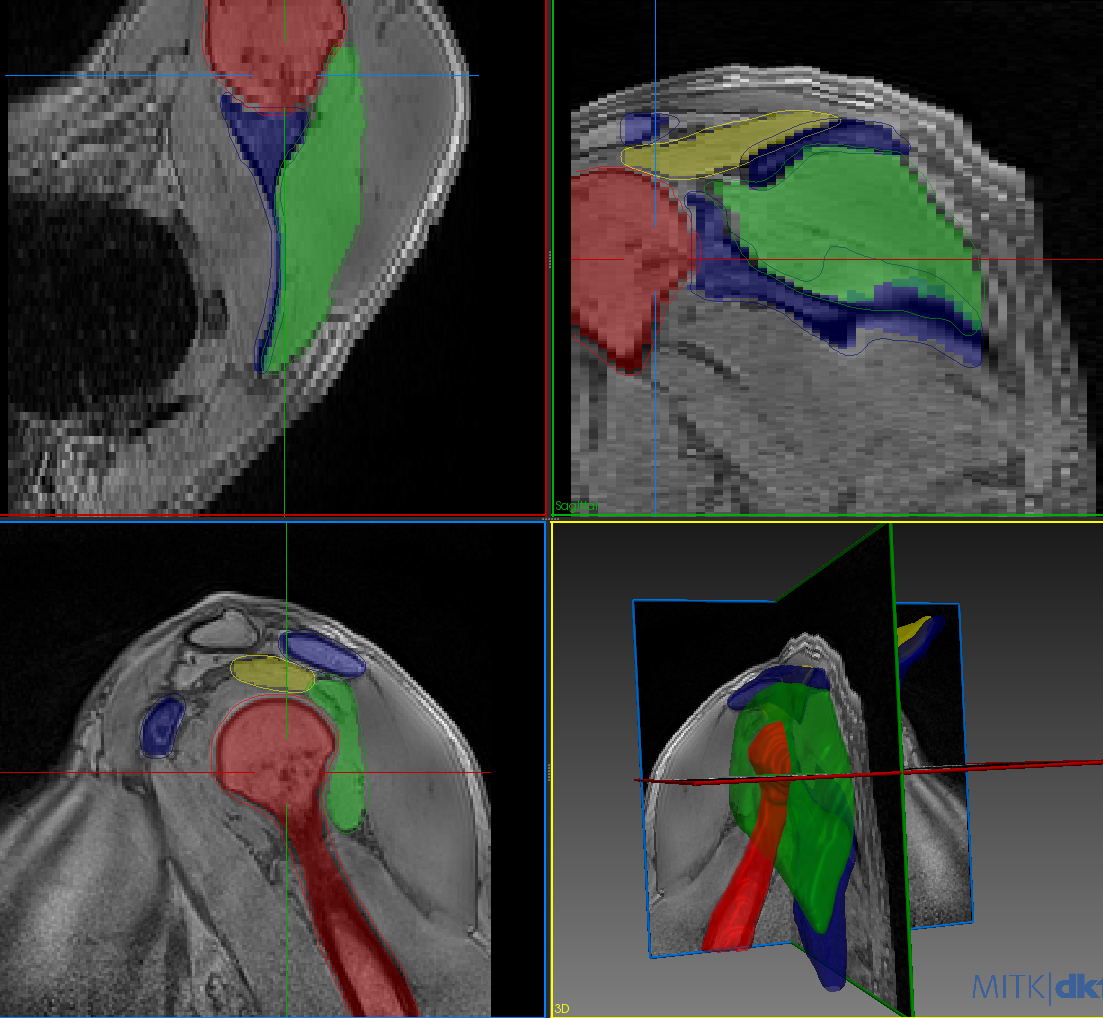}%
        \label{fig:data_samples:1}
    \end{subfigure}%
    \hspace{0.1em}%
    \begin{subfigure}[b]{0.49\linewidth}
    \centering
        \includegraphics[width=0.999\textwidth, trim= 0cm 0.0cm 0cm 0.3cm, clip]{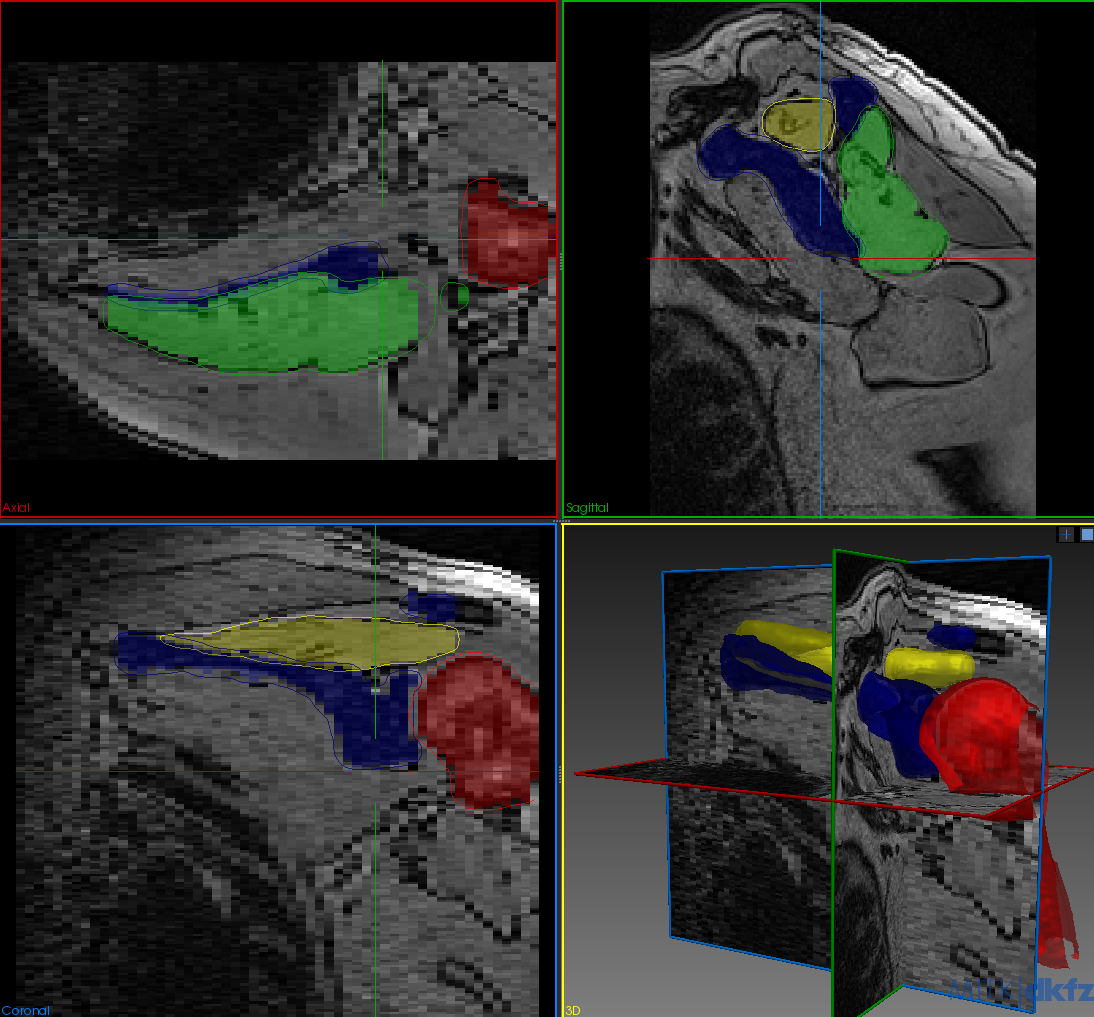}%
        \label{fig:data_samples:2}
    \end{subfigure}%
   	\caption{Cross-sectional view of 2 sample subject volumes along with expert annotations of humerus (red), scapula (blue), supraspinatus (yellow), and infraspinatus together with teres minor (green).
    }
    \label{fig:data_samples}
\end{figure}

\noindent
\textbf{Dataset.}
We have conducted experiments on an magnetic resonance imaging (MRI) dataset of 36 shoulders acquired with Dixon sequence with two slightly varying acquisition settings, resulting in the specifications shown in Table~\ref{tbl:dataset}.
For a more uniform dataset, images of the higher resolution setting \#2 were bilinearly interpolated to match the voxel resolution of the coarser dataset, and then zero padded to match the digital resolution of the images of that latter setting \#1.
The data has expert annotations of two bones (humerus \& scapula) and two muscle groups (supraspinatus \& infraspinatus + teres minor).
A cross-sectional view of two subjects along with the superimposed expert annotations are shown in Fig.~\ref{fig:data_samples}.
Ground truth annotations of setting \#2 were resized to match setting \#1 using nearest neighbor interpolation.
All experiments were conducted on the \textit{Total} dataset listed in Table~\ref{tbl:dataset}.

\vspace{1ex}\noindent
\textbf{Evaluation Metrics.}
For quantitative results, we evaluated Dice coefficient score and mean surface distance (MSD) metrics as commonly used metrics in medical image segmentation.
Dice score is $\mathrm{Dice}(M_S^l, M_G^l) = 2 |M_S^l \cap M_G^l| / (|M_S^l| + |M_G^l|)$, where $M_S^l$ is the binary predicted segmentation mask and $M_G^l$ the ground truth mask for label $l$.
MSD is computed symmetrically between the contours of segmentation prediction ($C_S$) and ground truth ($C_G$) for each label $l$ as, 
\begin{equation}
\mathrm{MSD}(C_S^l,C_G^l)  = \frac{\sum_{\mathrm{p} \in C_S^l} d(\mathrm{p}, C_G^l) + \sum_{\mathrm{p}' \in C_G^l} d(\mathrm{p}', C_S^l)}{|C_S^l| + |C_G^l|}
\end{equation}
where $d(\mathrm{p},C)$ is the closest Euclidean distance from point $\mathrm{p}$ to surface $C$. 
To compute the contour for a binary mask, we subtract its morphologically eroded version from itself using an erosion kernel of ${3\times3\times3}$.
Average Dice and MSD scores over the four given anatomical structures of interest are reported herein.

\vspace{1ex}\noindent
\textbf{Experimental Setup.}
Typically, expert annotations on MR volumes are conducted for all image slices of a volume at once when this is fetched manually from the PACS. 
However, this may lead to suboptimal use of limited annotation resources due to redundancy of annotating potentially similar images in a volume. 
A PACS compatible software can indeed fetch only the desired slices (2D images) from various volumes for annotation.
Therefore, we conducted experiments for both \textit{slice}-based and \textit{volume}-based active learning.
The former assumes the feasibility of random slice access and annotation query within $X_\mathrm{pool}$ whereas the latter treats each subject volume as an indivisible entity.

In an effort to efficiently utilize the available dataset, we generated 5 holdout sets using a pseudo random number generator where dataset splits were performed to roughly respect a $|D_\mathrm{pool}|$/validation/test ratio of $70\%/5\%/25\%$, with each subject being strictly in a single set. 
This yields to the following number of subjects: 25/2/9.
Then, slices of roughly one volume (i.e.,\, 64 slices for slice-based and all slices of a single subject for volume-based experiments) were randomly picked for each holdout set, to define the initial training set $D_\mathrm{an}^{(0)}$ and this initial set was kept constant across tests of different methods to ensure comparability. 

\section{Results}
\begin{figure}
\centering
	\begin{subfigure}[b]{0.509\linewidth}
    \centering
        \includegraphics[width=\textwidth, trim= 0cm 0.0cm 1cm 0.7cm, clip]{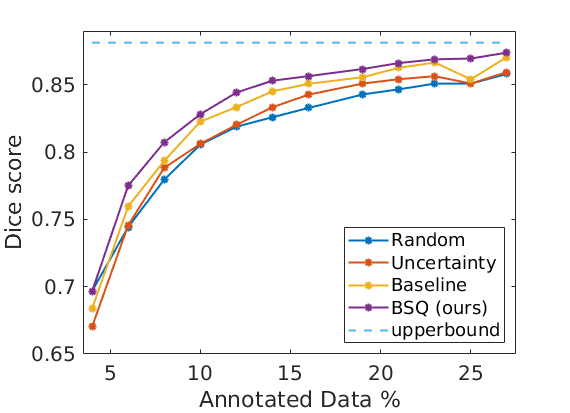}%
        \label{fig:scores_slice:dice}
    \end{subfigure}\hfill
    \begin{subfigure}[b]{0.489\linewidth}
    \centering
        \includegraphics[width=\textwidth, trim= 0cm 0.0cm 1cm 0.7cm, clip]{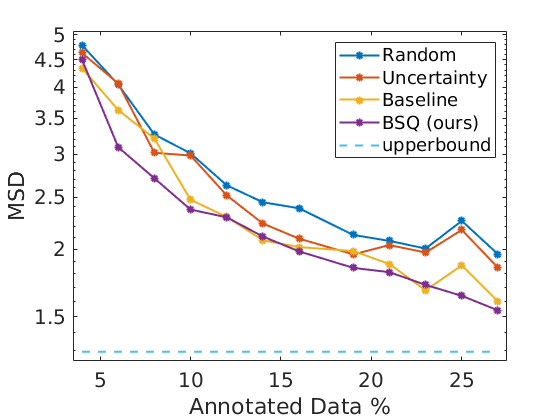}%
        \label{fig:scores_slice:assd}
    \end{subfigure}%
   	\caption{Average Dice score and mean surface distance (MSD) results of 2D image slice experiments for compared methods. 
    Upper bound marks the average performance when trained on entire $D_\mathrm{pool}$, i.e. the images of all 25 training volumes.
    }
    \label{fig:scores_slice}
\end{figure}

\begin{figure}
\centering
	\begin{subfigure}[b]{0.499\linewidth}
    \centering
        \includegraphics[width=\textwidth, trim= 0cm 0.0cm 1cm 0.7cm, clip]{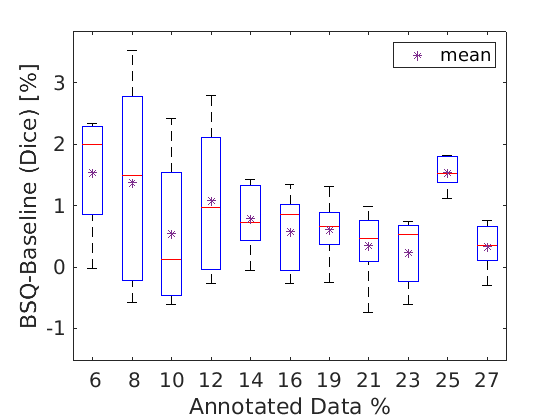}%
        \caption{2D Image Slices}
        \label{fig:dice_differences:slice}
    \end{subfigure}\hfill
    \begin{subfigure}[b]{0.499\linewidth}
    \centering
        \includegraphics[width=\textwidth, trim= 0cm 0.0cm 1cm 0.7cm, clip]{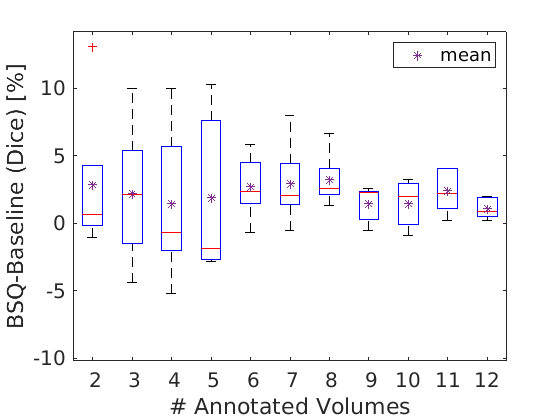}%
        \caption{3D Image Volumes}
        \label{fig:dice_differences:vol}
    \end{subfigure}%
   	\caption{
   	Dice score difference for each corresponding holdout set represented as box plot of their quartiles between the top two competing methods; $\mathrm{FCN}_\mathrm{BSQ}$ and $\mathrm{FCN}_\mathrm{Baseline}$.
   	Red lines show the median value, the blue boxes range from 25th to 75th percentiles, and purple stars show the mean values.
   	Overall positive values show superiority of $\mathrm{FCN}_\mathrm{BSQ}$ over $\mathrm{FCN}_\mathrm{Baseline}$, especially at earlier iterations.
   	}
    \label{fig:dice_differences}
\end{figure}

\noindent
\textbf{2D Image Slices.}
\label{sec:slice_based}
All slice-based experiments were initially trained on 64 slices.
For every active learning iteration, $n_\mathrm{unc}=64$ and $n_\mathrm{rep} = 32$ is used.
In Fig.~\ref{fig:scores_slice}, we show the Dice score and MSD of different methods over active learning iterations evaluated using the test set over 11 iterations, representing annotations from $4\%$ up to $27\%$ of the complete set $D_\mathrm{pool}$.
One can see that all compared methods achieve higher segmentation performance than randomly querying samples ($\mathrm{FCN}_\mathrm{Random}$). 
While the holdout set averages of Dice and MSD of $\mathrm{FCN}_\mathrm{Uncertainty}$ and $\mathrm{FCN}_\mathrm{Baseline}$ sometimes intersect, our proposed ($\mathrm{FCN}_\mathrm{BSQ}$) clearly outperforms all compared methods, shown as the purple curve in Fig.~\ref{fig:scores_slice}. 
To highlight the improvement that our proposed method brings over the baseline, we also present in Fig.~\ref{fig:dice_differences} the Dice score difference of the two methods with the highest quantitative performance, i.e.,\, $\mathrm{FCN}_\mathrm{BSQ}$ and $\mathrm{FCN}_\mathrm{Baseline}$. 
In Fig.~\ref{fig:dice_differences:slice} it can be observed that the Dice scores of $\mathrm{FCN}_\mathrm{BSQ}$ averaged over holdout sets are strictly superior to $\mathrm{FCN}_\mathrm{Baseline}$ at each presented iteration.
In Table~\ref{tbl:dice_diff_slice}, we list the mean and standard deviation of Dice score differences from the upper bound at different active learning iterations for the top two performing methods, $\mathrm{FCN}_\mathrm{BSQ}$ and $\mathrm{FCN}_\mathrm{Baseline}$.
Therein, one can see the percentage of the dataset that was annotated for these two methods in order to reach a segmentation performance within different tolerance limits from the upper bound. 

\begin{table*}
\centering
\caption{Dice score difference of $\mathrm{FCN}_\mathrm{BSQ}$ and $\mathrm{FCN}_\mathrm{Baseline}$ from upper bound for corresponding holdout set. 
Scores are from 2D Image Slice experiments and are presented as mean ($\pm$ standard deviation) [\%]. Lower values resemble closer performance to the upper bound.}
\label{tbl:dice_diff_slice}
\scriptsize
\setlength\tabcolsep{3pt}
\begin{tabular}{r|llllllllllll}
& \multicolumn{12}{c}{Annotation $\%$} \\
$\delta(\mathrm{Dice})\  [\%]$ & \multicolumn{1}{c}{4} & \multicolumn{1}{c}{6} & \multicolumn{1}{c}{8} & \multicolumn{1}{c}{10} & \multicolumn{1}{c}{12} & \multicolumn{1}{c}{14} & \multicolumn{1}{c}{16} & \multicolumn{1}{c}{19} & \multicolumn{1}{c}{21} & \multicolumn{1}{c}{23} & \multicolumn{1}{c}{25} & \multicolumn{1}{c}{27} \\ \hline
$\mathrm{FCN}_\mathrm{Baseline}$      & 19.8 (4.5) & 12.2 (2.7) & 8.8 (3.2) & 5.9 (2.8) & 4.8 (2.3) & 3.6 (1.8) & 3.1 (1.7) & 2.6 (1.6) & 1.9 (1.5) & 1.5 (1.6) & 2.7 (1.3) & 1.1 (1.6) \\
$\mathrm{FCN}_\mathrm{BSQ}$           & 18.5 (5.3) & 10.6 (2.6) & 7.4 (2.0) & 5.3 (1.7) & 3.7 (1.4) & 2.8 (1.3) & 2.5 (1.3) & 2.0 (1.2) & 1.5 (1.2) & 1.2 (1.1) & 1.2 (1.1) & 0.7 (1.3)
\end{tabular}
\end{table*}

\vspace{1ex}\noindent
\textbf{3D Image Volumes.}
\label{sec:volume_based}
In these experiments, the networks were initially trained on the slices of a single random subject ($D_\mathrm{an}^{(0)}$), and an active learning iteration consists of evaluating the respective scores of each method as an aggregation over a complete subject volume.
For $\mathrm{FCN}_\mathrm{BSQ}$ and $\mathrm{FCN}_\mathrm{Baseline}$, the set sizes of $S_\mathrm{unc}$ and $S_\mathrm{rep}$ are fixed to $n_\mathrm{unc}=2$ volumes and $n_\mathrm{rep} = 1$ volume.  

\begin{figure}
\centering
	\begin{subfigure}[b]{0.494\linewidth}
    \centering
        \includegraphics[width=\textwidth, trim= 0cm 0.0cm 1cm 0.7cm, clip]{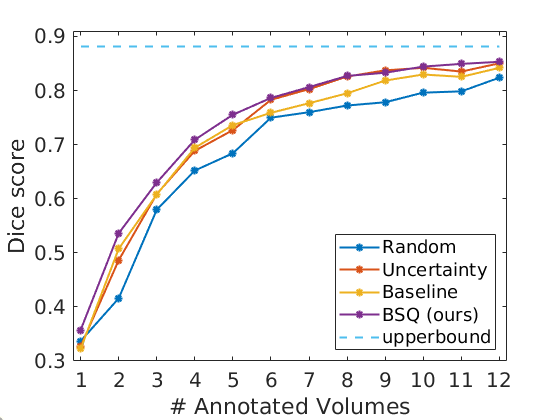}%
        \label{fig:scores_vol:dice}
    \end{subfigure}\hfill
    \begin{subfigure}[b]{0.504\linewidth}
    \centering
        \includegraphics[width=\textwidth, trim= 0cm 0.0cm 1.2cm 0.8cm, clip]{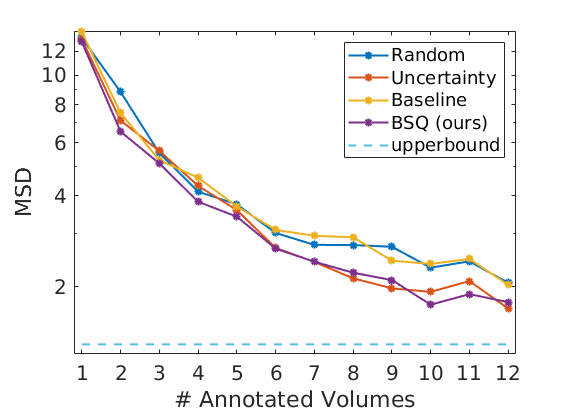}%
        \label{fig:scores_vol:assd}
    \end{subfigure}%
   	\caption{Average Dice score and mean surface distance (MSD) results of 3D volume experiments for compared methods. 
    Upper bound marks the average scores performance when trained on all 25 training volumes.
}
    \label{fig:scores_vol}
\end{figure}

Dice score and MSD of the compared methods for volume-based experiments are shown in Fig.~\ref{fig:scores_vol}.
Segmentation performance is evaluated at every active learning iteration, for a total of 11 iterations, using the same test set that was used for slice-based experiments. 
In the volume-based experiment results, where the annotation of entire volumes are added at each active learning iteration, the advantage of the compared methods appear more subtly, due to the larger range of Dice scores; e.g.,\, Dice scores ranging approximately from 0.3 to 0.9 as opposed to 0.65 to 0.9 in Fig.~\ref{fig:scores_slice}. 
Dice score improvement using $\mathrm{FCN}_\mathrm{BSQ}$ over $\mathrm{FCN}_\mathrm{Baseline}$ can be seen in Fig.~\ref{fig:dice_differences:vol}, where we show their Dice score differences between each holdout as boxplots.
One can see that $\mathrm{FCN}_\mathrm{BSQ}$ has improved average Dice score over $\mathrm{FCN}_\mathrm{Baseline}$ on every evaluation point (cf.~\ref{fig:dice_differences:vol} purple stars).
In order to have a precise understanding of the Dice score gap of the competing two methods from the upper bound, we present the mean and standard deviation of Dice score differences of $\mathrm{FCN}_\mathrm{BSQ}$ and $\mathrm{FCN}_\mathrm{Baseline}$ in Table~\ref{tbl:dice_diff_vol}.
\begin{table*}
\centering
\caption{Dice score difference of $\mathrm{FCN}_\mathrm{BSQ}$ and $\mathrm{FCN}_\mathrm{Baseline}$ from upper bound for corresponding holdout set. 
Scores are from 3D Image Volume experiments and are presented as mean ($\pm$ standard deviation) [\%]. Lower values resemble closer performance to the upper bound.}
\label{tbl:dice_diff_vol}
\scriptsize
\setlength\tabcolsep{2.5pt}
\begin{tabular}{r|rrrrrrrrrrrrllllllllllll}
& \multicolumn{12}{c}{\#Annotated volumes} \\
$\delta(\mathrm{Dice})\  [\%]$ & \multicolumn{1}{c}{1} & \multicolumn{1}{c}{2} & \multicolumn{1}{c}{3} & \multicolumn{1}{c}{4} & \multicolumn{1}{c}{5} & \multicolumn{1}{c}{6} & \multicolumn{1}{c}{7} & \multicolumn{1}{c}{8} & \multicolumn{1}{c}{9} & \multicolumn{1}{c}{10} & \multicolumn{1}{c}{11} & \multicolumn{1}{c}{12} \\ \hline
$\mathrm{FCN}_\mathrm{Baseline}$ & 55.9 (6.0) & 37.4 (4.1) & 27.4 (5.5) & 18.7 (8.7) & 14.5 (7.3) & 12.3 (5.1) & 10.5 (5.3) & 8.7 (2.8) & 6.3 (2.1) & 5.2 (1.8) & 5.6 (1.4) & 4.0 (1.4) \\
$\mathrm{FCN}_\mathrm{BSQ}$ & 52.4 (5.0) & 34.5 (2.1) & 25.2 (6.4) & 17.2 (4.1) & 12.6 (3.7) & 9.5 (4.1) & 7.5 (2.4) & 5.4 (1.9) & 4.9 (1.8) & 3.7 (1.6) & 3.2 (1.5) & 2.8 (0.9)
\end{tabular}
\end{table*}

\section{Discussion}
\label{sec:discussions}

Our preliminary experiments with VAE compared to infoVAE corroborate the claims in~\cite{zhao2017infovae} where the variance of the latent space was overestimated. 
Furthermore, active learning of segmentation through Bayesian sample querying using the above-mentioned VAE network trained on $X_\mathrm{pool}$ showed lower performance compared to $\mathrm{FCN}_\mathrm{BSQ}$.
Since we are strictly interested in the representational power of the latent variables for a given image, poorer performance on active learning evaluations indirectly support the claim of ``learning un-informative latent variables'' when using KL divergence~\cite{zhao2017infovae}. 

The advantage of $\mathrm{FCN}_\mathrm{Baseline}$ over $\mathrm{FCN}_\mathrm{Uncertainty}$ only becomes evident after a sufficiently large $D_\mathrm{an}$ is achieved ($\sim$$10\%$ in Fig.~\ref{fig:scores_slice}). 
One can also draw a similar conclusion by looking at Fig.~\ref{fig:dice_differences:slice}, where the superiority of our proposed method is most prominent in early iterations of active learning, and it almost monotonically decreases over time. 
This is inline with our previous findings in~\cite{ozdemir2018active} that an image descriptor based representativeness metric for $S_\mathrm{rep}$ may be redundant, if not adverse, until an adequate portion of the complete set is annotated.

All methods for the 3D volume experiment approach upper bound at a slower rate compared to the slice-based experiment (cf. Tables~\ref{tbl:dice_diff_slice}~\&~\ref{tbl:dice_diff_vol}).
This can be due to having less options to select from (i.e.,\, total of 24 volumes at first active learning iteration) when compared to slice-based. 
This hypothesis would be in line with the reasonable expectation that certain slices have significant importance for the segmentation task while others (e.g.,\, at the borders of the field-of-view) are less important in a given volume, whereas when a volume is given as a whole, the utilizable information therein is more uniform. 
Another point of interest is that after 6 volumes, $\mathrm{FCN}_\mathrm{Uncertainty}$ achieves performance closer to $\mathrm{FCN}_\mathrm{BSQ}$. 
This can possibly be due to the fact that our dataset consisted of 2 different settings (cf.\ Table~\ref{tbl:dataset}); where $\mathrm{FCN}_\mathrm{BSQ}$ may have early on queried for key sample volumes from both settings to represent the Total dataset, while $\mathrm{FCN}_\mathrm{Uncertainty}$ may have eventually seen key samples only after roughly 5 iterations of active learning. 
Another explanation can come from the design choice of assigning $n_\mathrm{unc}=2$ volumes and $n_\mathrm{rep} = 1$ volume, heavily restricting the sequential representativeness metric to pick one of the two options.

Upon comparison of the slice-based versus volume-based experimental setups, one can see the importance of querying slices as opposed to full volumes (e.g.,\, Dice score gap from the upper bound in early iterations of active learning on Tables~\ref{tbl:dice_diff_slice}~\&~\ref{tbl:dice_diff_vol}), i.e.,\, achieving better outcomes with less effort from experts.
Furthermore, a Dice score gap of approximately $5\%$ from the upper bound is achieved with $\mathrm{FCN}_\mathrm{BSQ}$ after merely 8 volumes ($32\%$ of $D_\mathrm{pool}$) in volume-based experiments, whereas a similar score is reached as early as $10\%$ for slice-based experiments.
In slice-based experiments, this gap drops to $2\%$ when less than a fifth ($19\%$) of the images are annotated; which equals to only $\sim$$285$ 2D image annotations, yielding a sufficiently high performance, compared to $\sim$$1500$ annotations necessitated for the upper bound scenario.

\section{Conclusions}
\label{sec:conclusions}

In this work, we have proposed a novel method to quantify representativeness of a sample from a large unsupervised dataset using Bayesian inference in the latent space of MMD VAEs.
We have shown that by using a learned mapping function onto a simple latent space and sample selection to align probability distributions in this space, the representational power of a subset of samples approach to that of the complete set, for the complex case of MR imaging.

Our results support the proposed approach being a suitable candidate for sample querying for the segmentation task in active learning.
Although our experimental dataset already harbors domain variation from two different acquisition settings, additional diversity is common in the clinical setting.
Consequently, the advantage of our proposed sample picking approach is expected to be more pronounced, by achieving a good coverage of the complete pool of images with only few active learning iterations and annotations.  
The main hypothesis herein is the representability of a dataset in the latent space as a continuous and Gaussian distribution. 
Future work shall investigate other means of dataset representations. 

\begingroup 
\let\thefootnote\relax\footnotetext{This work was funded by the Swiss National Science Foundation (SNSF) and a Highly Specialized Medicine (HSM2) grant of the Canton of Zurich.
We thank NVIDIA for their GPU support.} 
\endgroup

\bibliographystyle{model2-names}
\bibliography{bib}

\begin{thebibliography}{31}
\expandafter\ifx\csname natexlab\endcsname\relax\def\natexlab#1{#1}\fi
\providecommand{\url}[1]{\texttt{#1}}
\providecommand{\href}[2]{#2}
\providecommand{\path}[1]{#1}
\providecommand{\DOIprefix}{doi:}
\providecommand{\ArXivprefix}{arXiv:}
\providecommand{\URLprefix}{URL: }
\providecommand{\Pubmedprefix}{pmid:}
\providecommand{\doi}[1]{\href{http://dx.doi.org/#1}{\path{#1}}}
\providecommand{\Pubmed}[1]{\href{pmid:#1}{\path{#1}}}
\providecommand{\bibinfo}[2]{#2}
\ifx\xfnm\relax \def\xfnm[#1]{\unskip,\space#1}\fi
\bibitem[{Bengio et~al.(2007)Bengio, Lamblin, Popovici and
  Larochelle}]{bengio2007greedy}
\bibinfo{author}{Bengio, Y.}, \bibinfo{author}{Lamblin, P.},
  \bibinfo{author}{Popovici, D.}, \bibinfo{author}{Larochelle, H.},
  \bibinfo{year}{2007}.
\newblock \bibinfo{title}{Greedy layer-wise training of deep networks}, in:
  \bibinfo{booktitle}{Int Conf on Neural Information Processing Systems
  (NeurIPS)}, pp. \bibinfo{pages}{153--160}.
\bibitem[{Bromiley(2003)}]{bromiley2003products}
\bibinfo{author}{Bromiley, P.}, \bibinfo{year}{2003}.
\newblock \bibinfo{title}{Products and convolutions of gaussian probability
  density functions}.
\bibitem[{Chen et~al.(2016)Chen, Qi, Yu, Dou, Qin and Heng}]{Chen2016DCAN}
\bibinfo{author}{Chen, H.}, \bibinfo{author}{Qi, X.}, \bibinfo{author}{Yu, L.},
  \bibinfo{author}{Dou, Q.}, \bibinfo{author}{Qin, J.}, \bibinfo{author}{Heng,
  P.A.}, \bibinfo{year}{2016}.
\newblock \bibinfo{title}{{DCAN}: Deep contour-aware networks for object
  instance segmentation from histology images}.
\newblock \bibinfo{journal}{Medical Image Analysis} \bibinfo{volume}{36},
  \bibinfo{pages}{135--146}.
\bibitem[{Cohen et~al.(2017)Cohen, Afshar, Tapson and van
  Schaik}]{cohen2017emnist}
\bibinfo{author}{Cohen, G.}, \bibinfo{author}{Afshar, S.},
  \bibinfo{author}{Tapson, J.}, \bibinfo{author}{van Schaik, A.},
  \bibinfo{year}{2017}.
\newblock \bibinfo{title}{Emnist: Extending mnist to handwritten letters}, in:
  \bibinfo{booktitle}{IEEE Int Joint Conf on Neural Networks (IJCNN)}, pp.
  \bibinfo{pages}{2921--2926}.
\bibitem[{Feige(1998)}]{Feige1998a}
\bibinfo{author}{Feige, U.}, \bibinfo{year}{1998}.
\newblock \bibinfo{title}{A threshold of ln n for approximating set cover}.
\newblock \bibinfo{journal}{J. ACM} \bibinfo{volume}{45},
  \bibinfo{pages}{634--652}.
\bibitem[{Fran{\c{c}}ois(2008)}]{franccois2008high}
\bibinfo{author}{Fran{\c{c}}ois, D.}, \bibinfo{year}{2008}.
\newblock \bibinfo{title}{High-dimensional data analysis}, in:
  \bibinfo{booktitle}{From Optimal Metric to Feature Selection}.
  \bibinfo{publisher}{VDM Verlag Saarbrucken, Germany}, pp.
  \bibinfo{pages}{54--55}.
\bibitem[{F{\"u}rnstahl et~al.(2016)F{\"u}rnstahl, Schweizer, Graf,
  Vlachopoulos, Fucentese, Wirth, Nagy, Szekely and
  Goksel}]{furnstahl2016surgical}
\bibinfo{author}{F{\"u}rnstahl, P.}, \bibinfo{author}{Schweizer, A.},
  \bibinfo{author}{Graf, M.}, \bibinfo{author}{Vlachopoulos, L.},
  \bibinfo{author}{Fucentese, S.}, \bibinfo{author}{Wirth, S.},
  \bibinfo{author}{Nagy, L.}, \bibinfo{author}{Szekely, G.},
  \bibinfo{author}{Goksel, O.}, \bibinfo{year}{2016}.
\newblock \bibinfo{title}{Surgical treatment of long-bone deformities: {3D}
  preoperative planning and patient-specific instrumentation}, in:
  \bibinfo{booktitle}{Computational radiology for orthopaedic interventions}.
  \bibinfo{publisher}{Springer}, pp. \bibinfo{pages}{123--149}.
\bibitem[{Gal and Ghahramani(2016)}]{gal2016dropout}
\bibinfo{author}{Gal, Y.}, \bibinfo{author}{Ghahramani, Z.},
  \bibinfo{year}{2016}.
\newblock \bibinfo{title}{Dropout as a bayesian approximation: Representing
  model uncertainty in deep learning}, in: \bibinfo{booktitle}{Int Conf on
  Machine Learning (ICML)}, pp. \bibinfo{pages}{1050--1059}.
\bibitem[{Gal et~al.(2017)Gal, Islam and Ghahramani}]{gal2017deep}
\bibinfo{author}{Gal, Y.}, \bibinfo{author}{Islam, R.},
  \bibinfo{author}{Ghahramani, Z.}, \bibinfo{year}{2017}.
\newblock \bibinfo{title}{Deep bayesian active learning with image data}, in:
  \bibinfo{booktitle}{Int Conference on Machine Learning}, pp.
  \bibinfo{pages}{1183--1192}.
\bibitem[{Gretton et~al.(2006)Gretton, Borgwardt, Rasch, Sch\"{o}lkopf and
  Smola}]{Gretton2006a}
\bibinfo{author}{Gretton, A.}, \bibinfo{author}{Borgwardt, K.M.},
  \bibinfo{author}{Rasch, M.}, \bibinfo{author}{Sch\"{o}lkopf, B.},
  \bibinfo{author}{Smola, A.J.}, \bibinfo{year}{2006}.
\newblock \bibinfo{title}{A kernel method for the two-sample-problem}, in:
  \bibinfo{booktitle}{Int Conf on Neural Information Processing Systems
  (NeurIPS)}, pp. \bibinfo{pages}{513--520}.
\bibitem[{Guo et~al.(2019)Guo, Wang, Kang, Zhang, Gao and Li}]{guo2019btsdsn}
\bibinfo{author}{Guo, S.}, \bibinfo{author}{Wang, K.}, \bibinfo{author}{Kang,
  H.}, \bibinfo{author}{Zhang, Y.}, \bibinfo{author}{Gao, Y.},
  \bibinfo{author}{Li, T.}, \bibinfo{year}{2019}.
\newblock \bibinfo{title}{{BTS-DSN}: Deeply supervised neural network with
  short connections for retinal vessel segmentation}.
\newblock \bibinfo{journal}{International Journal of Medical Informatics}
  \bibinfo{volume}{126}, \bibinfo{pages}{105–113}.
\bibitem[{Kingma and Welling(2013)}]{Kingma2013autoencoding}
\bibinfo{author}{Kingma, D.P.}, \bibinfo{author}{Welling, M.},
  \bibinfo{year}{2013}.
\newblock \bibinfo{title}{Auto-encoding variational bayes}.
\newblock \bibinfo{journal}{arXiv preprint:1312.6114} .
\bibitem[{Konyushkova et~al.(2019)Konyushkova, Sznitman and
  Fua}]{konyushkova2019geometry}
\bibinfo{author}{Konyushkova, K.}, \bibinfo{author}{Sznitman, R.},
  \bibinfo{author}{Fua, P.}, \bibinfo{year}{2019}.
\newblock \bibinfo{title}{Geometry in active learning for binary and
  multi-class image segmentation}.
\newblock \bibinfo{journal}{Computer Vision and Image Understanding}
  \bibinfo{volume}{182}, \bibinfo{pages}{1--16}.
\bibitem[{LeCun et~al.(1998)LeCun, Bottou, Bengio, Haffner
  et~al.}]{lecun1998gradient}
\bibinfo{author}{LeCun, Y.}, \bibinfo{author}{Bottou, L.},
  \bibinfo{author}{Bengio, Y.}, \bibinfo{author}{Haffner, P.}, et~al.,
  \bibinfo{year}{1998}.
\newblock \bibinfo{title}{Gradient-based learning applied to document
  recognition}.
\newblock \bibinfo{journal}{Proceedings of the IEEE} \bibinfo{volume}{86},
  \bibinfo{pages}{2278--2324}.
\bibitem[{Lee et~al.(2015)Lee, Xie, Gallagher, Zhang and Tu}]{lee2015deeply}
\bibinfo{author}{Lee, C.Y.}, \bibinfo{author}{Xie, S.},
  \bibinfo{author}{Gallagher, P.}, \bibinfo{author}{Zhang, Z.},
  \bibinfo{author}{Tu, Z.}, \bibinfo{year}{2015}.
\newblock \bibinfo{title}{Deeply-supervised nets}, in:
  \bibinfo{booktitle}{Artificial intelligence and statistics}, pp.
  \bibinfo{pages}{562--570}.
\bibitem[{Li et~al.(2015)Li, Swersky and Zemel}]{Li2015generative}
\bibinfo{author}{Li, Y.}, \bibinfo{author}{Swersky, K.},
  \bibinfo{author}{Zemel, R.}, \bibinfo{year}{2015}.
\newblock \bibinfo{title}{Generative moment matching networks}, in:
  \bibinfo{booktitle}{Int Conf on Machine Learning (ICML)}, pp.
  \bibinfo{pages}{1718--1727}.
\bibitem[{Matthias et~al.(2018)Matthias, Karsten and Hanno}]{matthias2018deep}
\bibinfo{author}{Matthias, R.}, \bibinfo{author}{Karsten, K.},
  \bibinfo{author}{Hanno, G.}, \bibinfo{year}{2018}.
\newblock \bibinfo{title}{Deep bayesian active semi-supervised learning}, in:
  \bibinfo{booktitle}{IEEE Int Conference on Machine Learning and Applications
  (ICMLA)}, \bibinfo{organization}{IEEE}. pp. \bibinfo{pages}{158--164}.
\bibitem[{Milletari et~al.(2016)Milletari, Navab and
  Ahmadi}]{milletari2016vnet}
\bibinfo{author}{Milletari, F.}, \bibinfo{author}{Navab, N.},
  \bibinfo{author}{Ahmadi, S.A.}, \bibinfo{year}{2016}.
\newblock \bibinfo{title}{{V-Net}: Fully convolutional neural networks for
  volumetric medical image segmentation}, in: \bibinfo{booktitle}{IEEE Int Conf
  on 3D Vision (3DV)}, pp. \bibinfo{pages}{565--571}.
\bibitem[{Ozdemir et~al.(2018a)Ozdemir, Fuernstahl and
  Goksel}]{ozdemir2018learn}
\bibinfo{author}{Ozdemir, F.}, \bibinfo{author}{Fuernstahl, P.},
  \bibinfo{author}{Goksel, O.}, \bibinfo{year}{2018}a.
\newblock \bibinfo{title}{Learn the new, keep the old: Extending pretrained
  models with new anatomy and images}, in: \bibinfo{booktitle}{Medical Image
  Computing and Computer Assisted Intervention (MICCAI)}, pp.
  \bibinfo{pages}{361--369}.
\bibitem[{Ozdemir et~al.(2018b)Ozdemir, Peng, Tanner, Fuernstahl and
  Goksel}]{ozdemir2018active}
\bibinfo{author}{Ozdemir, F.}, \bibinfo{author}{Peng, Z.},
  \bibinfo{author}{Tanner, C.}, \bibinfo{author}{Fuernstahl, P.},
  \bibinfo{author}{Goksel, O.}, \bibinfo{year}{2018}b.
\newblock \bibinfo{title}{Active learning for segmentation by optimizing
  content information for maximal entropy}, in: \bibinfo{booktitle}{MICCAI
  Workshop on Deep Learning in Medical Image Analysis}.
  \bibinfo{publisher}{Springer}, pp. \bibinfo{pages}{183--191}.
\bibitem[{P{\'e}an et~al.(2017)P{\'e}an, Carrillo, F{\"u}rnstahl and
  Goksel}]{pean2017physical}
\bibinfo{author}{P{\'e}an, F.}, \bibinfo{author}{Carrillo, F.},
  \bibinfo{author}{F{\"u}rnstahl, P.}, \bibinfo{author}{Goksel, O.},
  \bibinfo{year}{2017}.
\newblock \bibinfo{title}{Physical simulation of the interosseous ligaments
  during forearm rotation}.
\newblock \bibinfo{journal}{EPiC Series in Health Sciences}
  \bibinfo{volume}{1}, \bibinfo{pages}{181--188}.
\bibitem[{Ronneberger et~al.(2015)Ronneberger, Fischer and
  Brox}]{ronneberger2015unet}
\bibinfo{author}{Ronneberger, O.}, \bibinfo{author}{Fischer, P.},
  \bibinfo{author}{Brox, T.}, \bibinfo{year}{2015}.
\newblock \bibinfo{title}{U-net: Convolutional networks for biomedical image
  segmentation}, in: \bibinfo{booktitle}{Medical Image Computing and Computer
  Assisted Intervention (MICCAI)}, pp. \bibinfo{pages}{234--241}.
\bibitem[{Salehi et~al.(2017)Salehi, Prevost, Moctezuma, Navab and
  Wein}]{Salehi2017precise}
\bibinfo{author}{Salehi, M.}, \bibinfo{author}{Prevost, R.},
  \bibinfo{author}{Moctezuma, J.L.}, \bibinfo{author}{Navab, N.},
  \bibinfo{author}{Wein, W.}, \bibinfo{year}{2017}.
\newblock \bibinfo{title}{Precise ultrasound bone registration with
  learning-based segmentation and speed of sound calibration}, in:
  \bibinfo{booktitle}{Medical Image Computing and Computer Assisted
  Intervention (MICCAI)}, pp. \bibinfo{pages}{682--690}.
\bibitem[{Sener and Savarese(2017a)}]{sener2017active}
\bibinfo{author}{Sener, O.}, \bibinfo{author}{Savarese, S.},
  \bibinfo{year}{2017}a.
\newblock \bibinfo{title}{Active learning for convolutional neural networks: A
  core-set approach}.
\newblock \bibinfo{journal}{arXiv preprint:1708.00489} .
\bibitem[{Sener and Savarese(2017b)}]{sener2017geometric}
\bibinfo{author}{Sener, O.}, \bibinfo{author}{Savarese, S.},
  \bibinfo{year}{2017}b.
\newblock \bibinfo{title}{A geometric approach to active learning for
  convolutional neural networks}.
\newblock \bibinfo{journal}{arXiv preprint:1708.00489v1} .
\bibitem[{Sinha et~al.(2019)Sinha, Ebrahimi and Darrell}]{sinha2019variational}
\bibinfo{author}{Sinha, S.}, \bibinfo{author}{Ebrahimi, S.},
  \bibinfo{author}{Darrell, T.}, \bibinfo{year}{2019}.
\newblock \bibinfo{title}{Variational adversarial active learning}, in:
  \bibinfo{booktitle}{IEEE Int Conf on Computer Vision (ICCV)}.
\bibitem[{Sirinukunwattana et~al.(2017)Sirinukunwattana, Pluim, Chen, Qi and
  Heng}]{Sirinukunwattana2017gland}
\bibinfo{author}{Sirinukunwattana, K.}, \bibinfo{author}{Pluim, J.P.},
  \bibinfo{author}{Chen, H.}, \bibinfo{author}{Qi, X.}, \bibinfo{author}{Heng,
  P.A.}, \bibinfo{year}{2017}.
\newblock \bibinfo{title}{Gland segmentation in colon histology images: The
  glas challenge contest}.
\newblock \bibinfo{journal}{Medical Image Analysis} \bibinfo{volume}{35},
  \bibinfo{pages}{489 -- 502}.
\bibitem[{Tompson et~al.(2015)Tompson, Goroshin, Jain, LeCun and
  Bregler}]{tompson2015efficient}
\bibinfo{author}{Tompson, J.}, \bibinfo{author}{Goroshin, R.},
  \bibinfo{author}{Jain, A.}, \bibinfo{author}{LeCun, Y.},
  \bibinfo{author}{Bregler, C.}, \bibinfo{year}{2015}.
\newblock \bibinfo{title}{Efficient object localization using convolutional
  networks}, in: \bibinfo{booktitle}{IEEE Conf on Computer Vision and Pattern
  Recognition (CVPR)}, pp. \bibinfo{pages}{648--656}.
\bibitem[{Yang et~al.(2017)Yang, Zhang, Chen, Zhang and
  Chen}]{yang2017suggestive}
\bibinfo{author}{Yang, L.}, \bibinfo{author}{Zhang, Y.}, \bibinfo{author}{Chen,
  J.}, \bibinfo{author}{Zhang, S.}, \bibinfo{author}{Chen, D.Z.},
  \bibinfo{year}{2017}.
\newblock \bibinfo{title}{Suggestive annotation: A deep active learning
  framework for biomedical image segmentation}, in: \bibinfo{booktitle}{Medical
  Image Computing and Computer Assisted Intervention (MICCAI)}, pp.
  \bibinfo{pages}{399--407}.
\bibitem[{Zeng et~al.(2017)Zeng, Yang, Li, Yu, Heng and Zheng}]{Zeng20173dunet}
\bibinfo{author}{Zeng, G.}, \bibinfo{author}{Yang, X.}, \bibinfo{author}{Li,
  J.}, \bibinfo{author}{Yu, L.}, \bibinfo{author}{Heng, P.A.},
  \bibinfo{author}{Zheng, G.}, \bibinfo{year}{2017}.
\newblock \bibinfo{title}{{3D U-N}et with multi-level deep supervision: Fully
  automatic segmentation of proximal femur in {3D MR} images}, in:
  \bibinfo{booktitle}{Machine Learning in Medical Imaging (MLMI)}, pp.
  \bibinfo{pages}{274--282}.
\bibitem[{Zhao et~al.(2017)Zhao, Song and Ermon}]{zhao2017infovae}
\bibinfo{author}{Zhao, S.}, \bibinfo{author}{Song, J.}, \bibinfo{author}{Ermon,
  S.}, \bibinfo{year}{2017}.
\newblock \bibinfo{title}{Infovae: Information maximizing variational
  autoencoders}.
\newblock \bibinfo{journal}{arXiv preprint:1706.02262} .

\end{thebibliography}

\clearpage
\appendix
\normalsize

\section{Further on Bayesian Sample Querying}

In this section, we conduct additional experiments to both visualize and quantify representativeness power of our Bayesian Sample Querying (BSQ) approach.
Although this work has investigated active learning for segmentation, experiments on simpler image-level classification tasks can clearer convey the merits of BSQ.
For instance, there are datasets of handwritten digits (e.g.,\, MNIST~\cite{lecun1998gradient} with 10 classes) and additional upper- and lower-case letters (e.g.,\, EMNIST~\cite{cohen2017emnist} with 62 classes) that contain grayscale images.
With the assumption that each character has different representative attributes, one can observe the entropy change over the proportion of class labels for the queried samples. 
In the ideal case, the distribution over the proportion of class labels should become uniform within the annotated dataset over time, causing the entropy to increase.
However, the categorically defined class labels can only be used as an auxiliary measure, since visual attributes, e.g., number of strokes, are not equally distant between classes.
Furthermore, the degree of variation within samples of different digits and letters varies heavily.
Consequently, one can attempt to observe and interpret the learned latent space under different constraints such as initially class-imbalanced annotation sets. 

\subsection{Experiment Setup \& Results}

Assuming that a latent space can capture all necessary degrees of variation using a few dimensions, we train an infoVAE for two setups; (1) MNIST dataset ($s_1$, $60$\,k samples
) using $n_\mathrm{lat} \! = \! 5$ dimensional latent space and (2) MNIST and EMNIST dataset ($s_2$, $\approx \! 758$\,k samples
) using $n_\mathrm{lat} \! = \! 10$ dimensional latent space, with a simple and mostly convolutional architecture similar to~\cite{zhao2017infovae}.
Next, we conduct 5 experiments, each simulating another imbalanced draw of the initial set of annotated data.
Accordingly, $n_\mathrm{init} \! = \! 10$ samples are drawn from the dataset randomly with probability $p_c(x)$ depending on the class $c$, where the priors for 3 randomly selected classes are reduced in each experiment by an order of magnitude. 
This is followed by $n_\mathrm{iter} \! = \! 30$ iterations of $n_\mathrm{rep} \! = \! 10$ representative sample queries, where the queried sample indices are determined using Eq.\,(\ref{eqn:log_bayes_query_numeric}).
Note that there is no uncertainty measure, since we are not assessing a classifier performance.
Code will be publicly available\footnote{\url{https://github.com/firatozdemir/AL-BSQ}}.

\begin{figure}
\centering
	{\includegraphics[width=0.40\textwidth, trim= 0cm 0.0cm 0cm 0cm, clip]{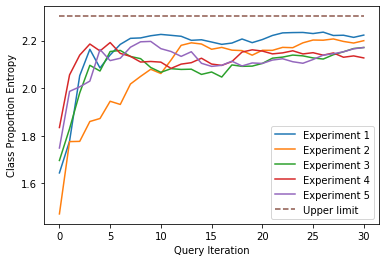}}%
	\\%
	{\includegraphics[width=0.40\textwidth, trim= 0cm 0.0cm 0cm 0cm, clip]{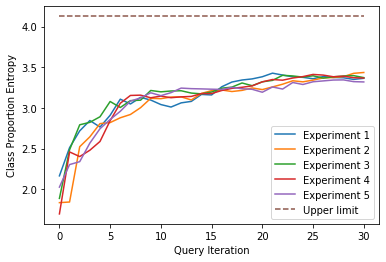}}%
    \caption{Entropy over the class proportion in the annotated dataset throughout 30 sample query iterations for setups (above) $s_1$ and (below) $s_2$.
    The upper limit is given by a uniform class proportions.
    }
    \label{fig:mnist_and_emnist_label_entropy}
\end{figure}

\noindent
\textbf{Class Entropy.}
As aforementioned, one can compute the entropy across classes; i.e.,\, $H(c) \! = \! -\sum_c^C p(c) \log p(c)$, where ${p(c)=1/M\sum_i^M \delta(y_i\!=\!c)}$, for the $M$ samples in the annotated dataset at each sample query iteration. 
Accordingly, the entropy values as new representative samples are queried for setups $s_1$ and $s_2$, respectively, are shown in Fig.~\ref{fig:mnist_and_emnist_label_entropy}.
It can be observed that for each experiment in both dataset setups, the entropy value, as expected, is steadily increasing over iterations as new samples are queried.

\begin{figure}
\centering%
\includegraphics[width=0.494\linewidth, trim= 0cm 0.0cm 0cm 0cm, clip]{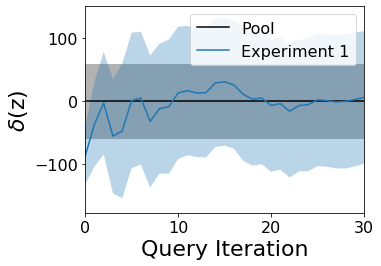}%
\hspace{0.1em}%
\includegraphics[width=0.494\linewidth, trim= 0cm 0.0cm 0cm 0cm, clip]{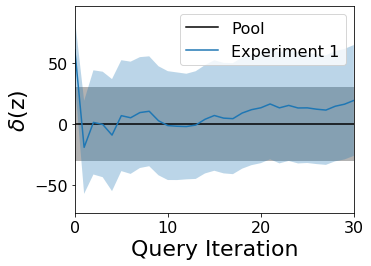}%
\\%
\includegraphics[width=0.494\linewidth, trim= 0cm 0.0cm 0cm 0cm, clip]{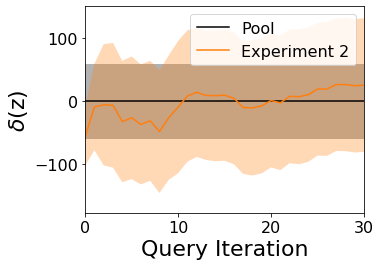}%
\hspace{0.1em}%
\includegraphics[width=0.494\linewidth, trim= 0cm 0.0cm 0cm 0cm, clip]{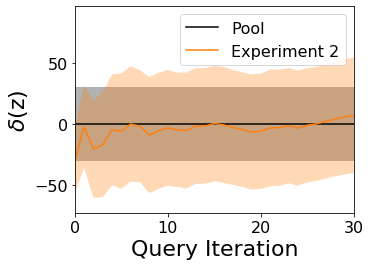}%
\\%
\includegraphics[width=0.494\linewidth, trim= 0cm 0.0cm 0cm 0cm, clip]{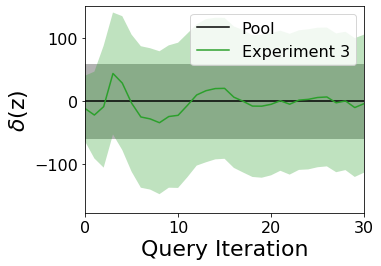}%
\hspace{0.1em}%
\includegraphics[width=0.494\linewidth, trim= 0cm 0.0cm 0cm 0cm, clip]{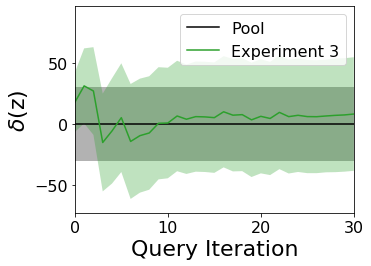}%
\\%
\includegraphics[width=0.494\linewidth, trim= 0cm 0.0cm 0cm 0cm, clip]{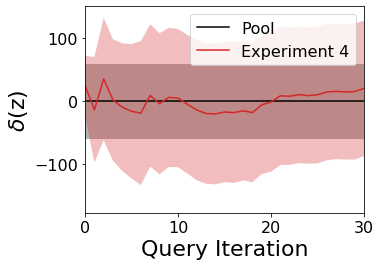}%
\hspace{0.1em}%
\includegraphics[width=0.494\linewidth, trim= 0cm 0.0cm 0cm 0cm, clip]{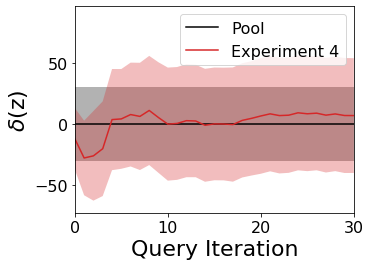}%
\\%
\includegraphics[width=0.494\linewidth, trim= 0cm 0.0cm 0cm 0cm, clip]{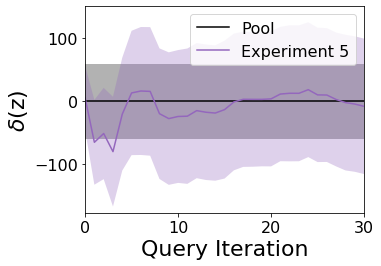}%
\hspace{0.1em}%
\includegraphics[width=0.494\linewidth, trim= 0cm 0.0cm 0cm 0cm, clip]{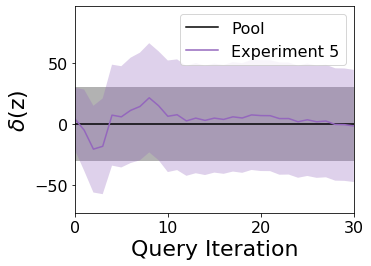}%
	\caption{
	The product of the fitted normal distributions on each latent dimension displayed using their mean and standard deviation for the (left) $s_1$ setup and (right) $s_2$ setup for 30 sample query iterations.
	$\delta(z)$ represents the axis shifted by the mean of the entire pool of images $\mu_{z^{*}_\mathrm{pool}}$, i.e.,\, $z-\mu_{z^{*}_\mathrm{pool}}$.
	Each row of image corresponds to an experiment, i.e.,\, a different random selection of initial annotated dataset.}
\label{fig:pdf_exps_30als_1d_joint}
\end{figure}

\noindent
\textbf{Latent Space Coverage.}
Alternatively, one can also observe the evolution of fitted normal distributions in the latent space. 
Let $q_{(\phi, s_i)}(x)$ be the learned mapping function onto the latent space $Z_\mathrm{s_i} \in \mathbb{R}^{n_\mathrm{lat}}$ of a corresponding experimental setup ($s_i$).
For a query iteration $t$, we map all samples from the annotated set $D_\mathrm{an}^{(t)}$ to the respective learned latent space and calculate the mean and standard deviation for each dimension to fit a multivariate diagonal Gaussian.
In order to qualitatively present the parameters of the fitted Gaussians in 1D, we treat each dimension of the multivariate Gaussian as a univariate Gaussian, and compute the parameters of the product of $n_\mathrm{lat}$ univariate Gaussians following~\cite{bromiley2003products}. 
In Fig.~\ref{fig:pdf_exps_30als_1d_joint}, the evolution of the fitted Gaussians are shown for all five experiments throughout 30 query iterations for both setups $s_1$~\&~$s_2$, where the y-axis is shifted to the mean value of the entire pool of images $\mu_{z^{*}_\mathrm{pool}}$.
It can be seen that the first few iterations counter the shifted mean of the imbalanced initial annotated set. 
The following iterations evolve around the mean of the $X_\mathrm{pool}$. 
Eq.\,(\ref{eqn:log_bayes_query_numeric}) promotes selecting samples away from already annotated ones, which encourages covering the same range of representative attributes as $X_\mathrm{pool}$ with substantially fewer samples.  
Consequently, $X_\mathrm{an}$ has a higher standard deviation in the representative space, which is also observed in Fig.~\ref{fig:pdf_exps_30als_1d_joint}.
These empirical results corroborate the hypothesis that Eq.\,(\ref{eqn:log_bayes_query_numeric}) promotes selecting samples that cover the mode and breadth of the distribution of $D_\mathrm{pool}$ in the representational space.

\end{document}